\documentclass[letterpaper, 10 pt, conference]{ieeeconf}  

\IEEEoverridecommandlockouts                              

\usepackage{graphicx} 
\usepackage{epsfig} 
\usepackage{mathptmx} 
\usepackage{times} 
\usepackage{amsmath} 
\usepackage{amssymb}  
\usepackage{booktabs}
\usepackage[ruled,vlined]{algorithm2e}  
\usepackage{xcolor}  

\usepackage{multirow}     
\usepackage{array}        
\usepackage{colortbl}     
\usepackage{caption}      
\usepackage{makecell}     
\usepackage{bm}           
\usepackage{arydshln}     
\usepackage{multirow}       
\usepackage{colortbl}       
\usepackage{xcolor}         
\usepackage{booktabs}       
\usepackage{graphicx}       
\usepackage{caption}        
\usepackage{fontsize}  
\newcommand{\thickhline}{%
    \noalign{\hrule height 0.8pt}
}

\newcolumntype{I}{!{\vrule width 1pt}}

\definecolor{RedOrange}{RGB}{255,83,73}

\definecolor{DarkBlue}{RGB}{0,0,139}
\definecolor{Goldenrod}{RGB}{218,165,32}
\definecolor{SkyBlue}{RGB}{135,206,235}

\SetKwComment{tcc}{/* }{ */}  
\DontPrintSemicolon  

\makeatletter
\let\NAT@parse\undefined
\makeatother
\usepackage[colorlinks=true, linkcolor=blue, citecolor=blue, urlcolor=blue]{hyperref}

\usepackage{cleveref}  

\Crefname{equation}{Eq.}{Eqs.}

\title{\LARGE \bf
ActiveSSF: An Active-Learning-Guided Self-Supervised Framework for Long-Tailed Megakaryocyte Classification
}

\author{Linghao Zhuang$^{1,*}$, 
    Ying Zhang$^{2,3,4,*}$, 
    Gege Yuan$^{5,*}$, 
    Xingyue Zhao$^{5}$ and
    Zhiping Jiang$^{2,3,4,\dagger}$
    \thanks{$^{}$ This research was supported by the Hunan Provincial Natural Science Foundation, China, under the General Project program (S2023JJMSXM1473).}
    \thanks{$^{*}$ These authors contributed equally to this work.}
    \thanks{$^{1}$ School of Software Engineering, Xinjiang University, Urumqi, China.}%
    \thanks{$^{2}$ Department of Hematology, Xiangya Hospital, Central South University, Changsha, China.}%
    \thanks{$^{3}$ National Clinical Research Center for Geriatric Diseases, Xiangya Hospital, Changsha, China.}%
    \thanks{$^{4}$ Hunan Hematology Oncology Clinical Medical Research Center, Changsha, China.}%
    \thanks{$^{5}$ School of Software Engineering, Xi'an Jiaotong University, Xi'an, China.}%
    \thanks{$^{\dagger}$ Corresponding author. Email address: Jiangzhp@csu.edu.cn}
}

\begin{document}

\maketitle
\thispagestyle{empty}
\pagestyle{empty}


\begin{abstract}

Precise classification of megakaryocytes is crucial for diagnosing myelodysplastic syndromes. Although self-supervised learning has shown promise in medical image analysis, its application to classifying megakaryocytes in stained slides faces three main challenges: (1) pervasive background noise that obscures cellular details, (2) a long-tailed distribution that limits data for rare subtypes, and (3) complex morphological variations leading to high intra-class variability. To address these issues, we propose the ActiveSSF framework, which integrates active learning with self-supervised pretraining. Specifically, our approach employs Gaussian filtering combined with K-means clustering and HSV analysis (augmented by clinical prior knowledge) for accurate region-of-interest extraction; an adaptive sample selection mechanism that dynamically adjusts similarity thresholds to mitigate class imbalance; and prototype clustering on labeled samples to overcome morphological complexity. Experimental results on clinical megakaryocyte datasets demonstrate that ActiveSSF not only achieves state-of-the-art performance but also significantly improves recognition accuracy for rare subtypes. Moreover, the integration of these advanced techniques further underscores the practical potential of ActiveSSF in clinical settings.



\end{abstract}

\section{INTRODUCTION}

Megakaryocytes are essential components of the hematological system, and their accurate classification is crucial for the early diagnosis of myelodysplastic syndrome (MDS)~\cite{su2014neural, TOMARI2014206}. In recent years, rapid advances in deep learning have dramatically enhanced both the accuracy and efficiency of computer-aided diagnosis in medical imaging~\cite{dong2022fusing, 10508987, 10635682, zhao2024hfgs, li2023d}. However, automatic classification of megakaryocytes still encounters several challenges, including the underutilization of vast amounts of unlabeled data, limited discriminative features in cell images, and imbalanced class distributions among megakaryocyte subtypes. To address these issues, various deep learning–based methods have been proposed for cell classification—for example, Su et al.~\cite{su2014neural} leveraged the HSI color space to extract key cellular features and combined them with neural networks for efficient classification.

\begin{figure}[tb]
    \vspace{5px}
    \centering
    \includegraphics[width=0.45\textwidth]{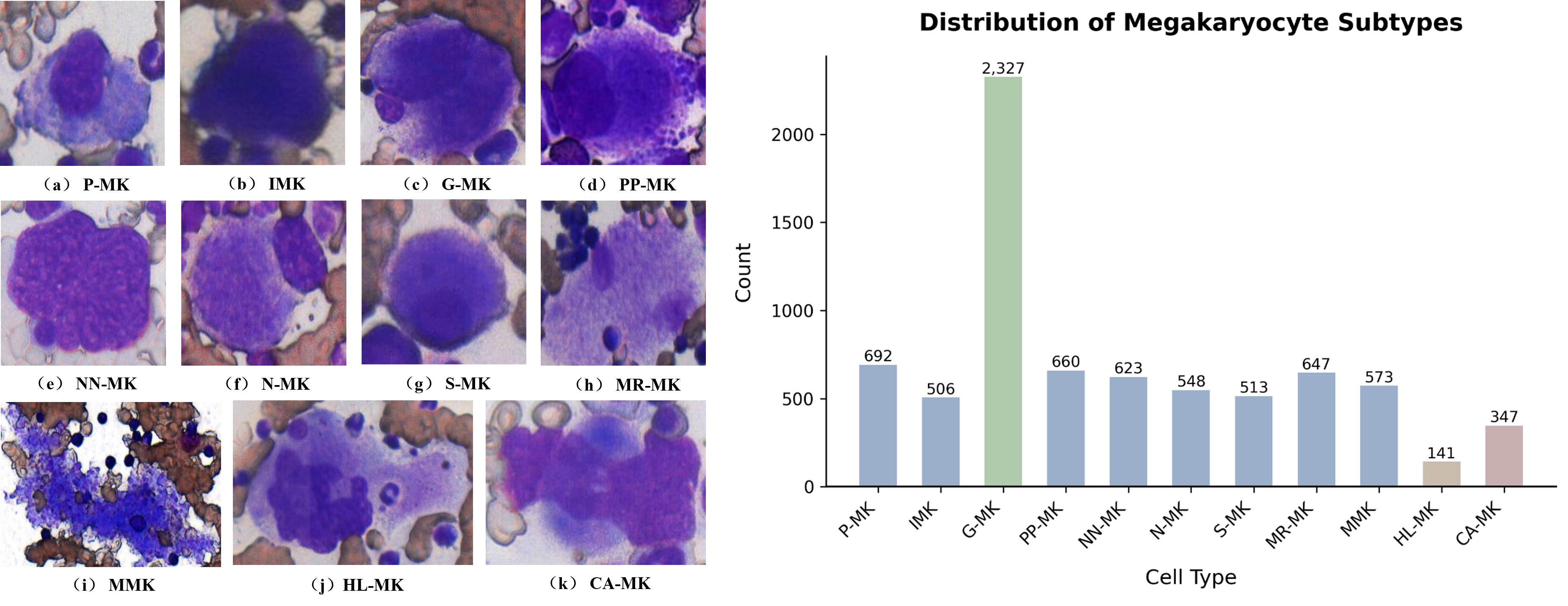}
    \vspace{-5px}
    \caption{
    Overview of our megakaryocyte dataset. Left: Representative images of different megakaryocyte subtypes. Right: Distribution of megakaryocyte subtypes showing the inherent class imbalance.
    }
    \label{fig:fig1}
    \vspace{-20px}
\end{figure}    

\begin{figure*}[!ht]
  \centering
  \vspace{10px}
  \includegraphics[width=0.9\textwidth]{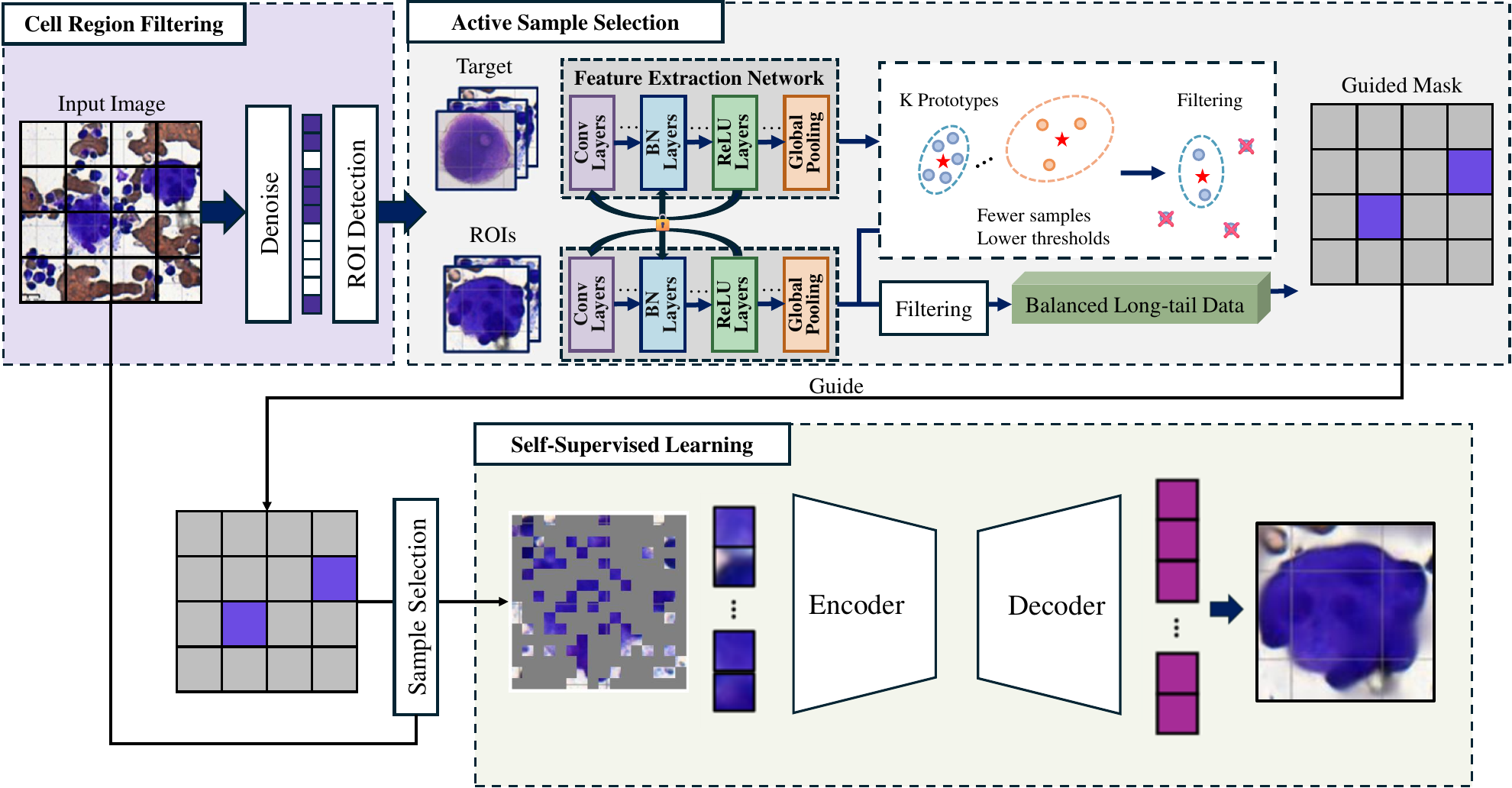}
  \caption{
  We present a two-stage active learning framework for guiding self-supervised pretraining. Stage 1 (Cell Region Filtering) applies Gaussian blur and K-means clustering to isolate cellular regions from background noise. Stage 2 (Active Sample Selection) extracts features via ResNet~\cite{he2016identity}, clusters them into K prototypes using K-means~\cite{macqueen1967some}, and establishes dynamic thresholds inversely proportional to cluster size—thereby accommodating rare subtypes with lower thresholds. The framework selects samples based on their distance to cluster centers, retaining only those within their respective thresholds for self-supervised pretraining.
  }
  \label{fig:Framework}
  \vspace{-15px}
\end{figure*}

Despite these advances, effectively utilizing unlabeled data remains a significant hurdle. Recently, self-supervised learning has emerged as a promising approach that enables models to learn robust representations from unlabeled data~\cite{chen2021empirical, gao2022convmae, Chen_2023_CVPR, he2022masked, he2020momentum}. For instance, the Masked Autoencoder (MAE) introduced by He et al.~\cite{he2022masked} reconstructs missing pixels by randomly masking regions in large-scale images, thereby capturing essential structural information. Nevertheless, given the inherently limited discriminative features in megakaryocyte images and the challenge of class imbalance, current self-supervised models continue to face difficulties in this classification task.

In this study, we present a comprehensive dataset comprising eleven distinct megakaryocyte subtypes (Fig.~\ref{fig:fig1}), enabling detailed characterization of their morphological diversity. The dataset includes: Primitive megakaryocyte (P-MK), Immature megakaryocyte (IMK), Megakaryocyte with cytoplasmic abnormalities (CA-MK), Granular megakaryocyte (G-MK), Platelet-producing megakaryocyte (PP-MK), Naked nucleus megakaryocyte (NN-MK), Normal-sized megakaryocyte with unlobed or minimally lobed nucleus (N-MK), Small megakaryocyte with unlobed or minimally lobed nucleus (S-MK), Micromegakaryocyte (mMK), Multinucleated round megakaryocyte (MR-MK), and Large megakaryocyte with highly lobulated nucleus (HL-MK). Analysis reveals a pronounced long-tailed distribution, where G-MK predominates while CA-MK and HL-MK occur rarely. To our knowledge, this represents the first application of deep learning techniques for fine-grained classification of megakaryocyte subtypes. Examination of routine bone marrow smears demonstrates that images not only contain various non-megakaryocyte cells but also extensive background regions, with certain subtypes appearing extremely infrequently. These observations motivated the development of ActiveSSF—a novel self-supervised learning framework that integrates active learning strategies during pretraining to enhance model performance on this challenging dataset.

Specifically, to mitigate the issue of extensive background regions in slide images, we first apply Gaussian blur and K-means clustering~\cite{macqueen1967some} to reduce noise, followed by focusing on high-energy density regions in the HSV color space to extract key cellular information—a process we term Cell Region Filtering. Furthermore, recognizing the complex morphological variations exhibited by megakaryocytes—where intra-class differences often surpass inter-class differences—we begin by clustering the labeled samples to generate representative prototypes for each subtype. These prototypes are subsequently compared with unlabeled samples to accurately isolate megakaryocytes while filtering out irrelevant ones. To effectively tackle the long-tailed distribution, we dynamically adjust similarity thresholds, applying higher thresholds for common subtypes and lower thresholds for rare subtypes—a process we refer to as Active Sample Selection.

Collectively, these active learning techniques guide the pretraining data selection process, significantly enhancing the utilization of unlabeled data and the recognition of rare samples. Experimental results demonstrate that our proposed method substantially improves classification accuracy—particularly for rare megakaryocyte subtypes—thereby underscoring its potential for clinical applications in MDS diagnosis.

\section{Methods}

\subsection{Overview of the Method}

As illustrated in Fig.~\ref{fig:Framework}, our framework comprises two key stages: cell region filtering and active sample selection. In the cell region filtering stage, we first apply a combination of Gaussian blur and K-means clustering~\cite{macqueen1967some} to suppress background noise, and then utilize HSV color space analysis along with clinical prior knowledge to accurately extract the regions corresponding to target cells. In the active sample selection stage, we extract features from labeled samples using a ResNet~\cite{he2016identity} backbone network, and perform clustering to generate representative prototypes. Notably, we propose a dynamic similarity threshold adjustment mechanism—where the threshold is modulated based on the size of each cluster—to guide the selection of samples from the unlabeled target domain. This adaptive strategy prioritizes samples from rare classes, thereby effectively mitigating the long-tailed distribution issue during self-supervised pretraining.

\subsection{Cell Region filtering based on clinical priors}

\subsubsection{Noise Suppression}
Megakaryocyte images often contain complex background noise and non-uniform illumination, which significantly impact classification accuracy. To address this challenge, we propose a preprocessing method based on Gaussian filtering and clustering. For each image \( D_i \) in the dataset \( D \), the processing pipeline proceeds as follows:

First, to mitigate variations in illumination, we normalize the image using standard normalization techniques. Subsequently, we apply a \( 3 \times 3 \) Gaussian kernel with \( \sigma = 1.5 \) for spatial filtering, as described by:

\begin{equation}
G(x, y) = \frac{1}{2\pi\sigma^2} e^{ -\frac{x^2 + y^2}{2\sigma^2} }.
\end{equation}

This filtering operation effectively suppresses high-frequency noise, generating a smoothed intermediate image \( G_i \). To further separate cellular regions from the background, we design a two-stage K-means clustering strategy~\cite{macqueen1967some}. In the first stage, the algorithm partitions the image into 20 color clusters to capture the primary color distribution patterns; in the second stage, adaptive cluster merging refines the number of clusters to 10. The cluster centers are updated as follows:

\begin{equation}
c_j = \frac{1}{|S_j|} \sum_{p_i \in S_j} p_i,
\end{equation}
where \( p_i \in \mathbb{R}^3 \) represents the RGB values of a pixel, and \( S_j \) denotes the set of pixels in cluster \( j \). The optimization objective is to minimize the squared Euclidean distance between pixels and their corresponding cluster centers:

\begin{equation}
d(p_i, c_j) = \| p_i - c_j \|^2_2.
\end{equation}

This iterative optimization continues until convergence, ultimately yielding the denoised image \( G'_i \). Our hierarchical processing strategy not only effectively suppresses background interference but also preserves key morphological features of megakaryocytes, thereby laying a solid foundation for subsequent classification tasks. Nevertheless, while noise suppression greatly improves image quality, accurately identifying megakaryocyte regions in these preprocessed images remains challenging due to their complex morphological characteristics. To address this, we further propose a specialized detection method leveraging the HSV color space.

\subsubsection{Cell Region Detection in HSV Color Space}

Accurate identification of megakaryocyte regions is crucial for subsequent classification tasks. However, the RGB color space often fails to effectively distinguish cells from the background. To address this limitation, we propose a region extraction method that combines HSV color space analysis with morphological feature filtering.

First, we transform the preprocessed image \( G' \) into the HSV color space to better capture the chromatic properties of the cells. Based on clinical expertise, we establish two specific HSV threshold ranges: one for a purple mask (H: 30--140, S: 100--255, V: 0--255) and another for a deep blue mask (e.g., H: 95--105, S: 150--255, V: 50--255) to capture the characteristic staining patterns of megakaryocytes. These thresholds were carefully determined through empirical validation with clinical experts. By performing a pixel-wise logical OR operation on these two masks, we obtain a composite mask that effectively highlights potential cellular regions while minimizing false positives from background artifacts.

To further enhance detection precision, we introduce a morphological feature-based filtering mechanism. For each detected region, we compute its bounding box and apply two key constraints: the bounding box must be at least \(70 \times 70\) pixels in size, and the region must have a fill rate of at least 70\% (i.e., at least 70\% of the pixels within the bounding box belong to the candidate region). This dual-constraint approach effectively filters out background noise and non-target regions. Consequently, for image \(i\), we obtain \(n\) high-quality candidate regions denoted as \( S_i = \{ S_1, \ldots, S_n \} \).

With reliable cell regions identified, the next crucial step is to develop an effective strategy for selecting representative samples, particularly for rare subtypes, to guide the self-supervised learning process.

\subsection{Active Sample Selection}

\subsubsection{Clustering on the Labeled Samples}

Effective prototype generation from labeled samples is crucial for guiding unlabeled data selection in self-supervised learning. However, directly applying clustering on raw features may lead to suboptimal prototypes due to the complex nature of megakaryocyte morphology. To address this challenge, we propose a feature-based clustering strategy that leverages deep neural networks and iterative optimization.

Specifically, we first employ the ResNet50~\cite{he2016identity} backbone network to extract high-dimensional features \( F = \{ f_1, \ldots, f_n \} \) from the labeled dataset \( T \). These features capture essential morphological characteristics of megakaryocytes. We then apply K-means clustering~\cite{macqueen1967some} with the number of clusters set to 11 to generate representative prototypes through iterative optimization. The objective function is defined as:

\begin{equation}
\mathcal{L} = \sum_{i=1}^{K} \sum_{x \in C_i} \| x - c_i \|^2_2,
\end{equation}
where \( c_i \) represents the center of cluster \( i \), and \( x \) denotes the feature vector of a sample. The algorithm updates cluster assignments based on:

\begin{equation}
C_i = \{ x : \| x - c_i \|^2_2 \leq \| x - c_j \|^2_2 \quad \forall j = 1, \ldots, K \},
\label{eq:cluster}
\end{equation}
and refines cluster centers through:

\begin{equation}
c_i = \frac{1}{|C_i|} \sum_{x \in C_i} x.
\label{eq:cluster1}
\end{equation}

This iterative process continues until convergence, yielding a set of robust prototypes that effectively characterize the different megakaryocyte subtypes. These prototypes serve as reliable references for subsequent sample selection, particularly benefiting the identification of rare subtypes in the unlabeled data pool. Building upon these robust prototypes, we further introduce an adaptive sample selection strategy to address the challenge of long-tailed distribution in megakaryocyte datasets.

\subsubsection{Sample Selection}

A critical challenge in megakaryocyte classification is the long-tailed distribution of cell subtypes, where rare subtypes often receive insufficient attention during training. To address this issue, we propose an adaptive sample selection strategy that dynamically adjusts the selection criteria based on class distribution.

For each candidate region \( S_i \) obtained from the previous cell detection stage, we first extract deep features using the ResNet~\cite{he2016identity} backbone network. The representativeness of each sample is then evaluated by computing its Euclidean distance to the nearest cluster prototype:

\begin{equation}
d(x, c) = \| x - c \|_2,
\end{equation}
where \( x \in \mathbb{R}^d \) denotes the extracted feature vector and \( c \) corresponds to its closest cluster center.

To balance the selection between common and rare subtypes, we introduce a density-aware threshold mechanism:

\begin{equation}
\text{threshold}_i = \text{LB} + (\text{UB} - \text{LB}) \times \left(1 - \frac{n_i}{n_{\max}} \right)^{\alpha},
\label{eq:threshold}
\end{equation}
where \(\text{LB}\) and \(\text{UB}\) are the minimum and maximum distances within each cluster, \( n_i \) is the number of samples in cluster \( i \), \( n_{\max} \) is the size of the largest cluster, and \(\alpha\) is a scaling factor that controls the adjustment rate. This adaptive mechanism automatically lowers the threshold for rare subtypes (\( n_i \ll n_{\max} \)) while maintaining stricter criteria for common subtypes (\( n_i \approx n_{\max} \)).

The final sample selection is performed as follows:

\begin{equation}
S'_i = \{ x \mid d(x, c_i) \leq \text{threshold}_i \},
\end{equation}
where \( S'_i \) denotes the set of selected samples for cluster \( i \). By integrating deep features with adaptive thresholds, this strategy effectively prioritizes the inclusion of underrepresented subtypes while ensuring quality control for common ones, thereby addressing the inherent class imbalance in megakaryocyte datasets.

Moreover, as outlined in Algorithm~\ref{alg:activessf}, our ActiveSSF framework seamlessly integrates cell region filtering, prototype generation, and adaptive sample selection into a unified pipeline, providing a systematic approach to leverage unlabeled data while mitigating the challenges posed by the long-tailed distribution in megakaryocyte classification.

\section{Experiment and Results}
\label{sec:typestyle}

\begin{table*}[ht]\small
\centering
\vspace{10px}
\scriptsize{
\resizebox{\linewidth}{!}{
\setlength\tabcolsep{3.pt}
\renewcommand\arraystretch{1.1}
\begin{tabular}{r||ccccccccccc}
\hline\thickhline
\rowcolor{gray!20}
Methods & P-MK & IMK & CA-MK$^{\dagger}$ & G-MK$^{*}$ & PP-MK & NN-MK & N-MK & S-MK & MMK & MR-MK & HL-MK$^{\dagger}$  \\
\hline\hline
ResNet~\cite{he2016identity}
& 98.02 ± 1.12 & 69.21 ± 4.84 & 64.76 ± 8.01 & 83.34 ± 1.64 & 68.52 ± 3.66 & 95.64 ± 1.94 & 64.92 ± 1.31 & 61.73 ± 5.86 & 95.74 ± 1.22 & 88.52 ± 4.82 & 65.18 ± 6.91
\\
ViT~\cite{alexey2020image}
& 98.15 ± 0.94 & 62.95 ± 8.89 & 52.91 ± 11.12 & 72.81 ± 2.48 & 58.42 ± 3.37 & 93.57 ± 0.93 & 51.54 ± 5.50 & 48.66 ± 6.58 & 92.24 ± 3.10 & 70.45 ± 2.64 & 43.85 ± 16.73
\\
\hdashline
MoCo v3~\cite{chen2021empirical} w/o ActiveSSF
& 97.51 ± 1.23 & 68.71 ± 5.16 & 57.62 ± 7.55 & 78.79 ± 2.19 & 65.31 ± 2.03 & 95.20 ± 0.57 & 55.57 ± 4.74 & 52.43 ± 8.55 & 94.14 ± 1.77 & 81.36 ± 4.32 & \textbf{59.75 ± 13.04}
\\
\rowcolor{gray!10} MoCo v3 w/ ActiveSSF
& \textbf{97.64 ± 1.38} & \textbf{71.85 ± 5.82} & \textbf{60.92 ± 6.82} & \textbf{79.70 ± 1.91} & \textbf{65.45 ± 3.37} & \textbf{96.17 ± 0.27} & \textbf{57.34 ± 4.59} & \textbf{57.62 ± 6.49} & \textbf{94.61 ± 1.80} & \textbf{85.45 ± 4.64} & 58.11 ± 10.00
\\
\hdashline
ConvMAE~\cite{gao2022convmae} w/o ActiveSSF
& 98.49 ± 1.09 & 74.33 ± 5.77 & 68.09 ± 7.63 & 83.98 ± 0.69 & 70.76 ± 2.78 & 96.99 ± 0.87 & 62.22 ± 4.15 & 64.06 ± 10.90 & 95.75 ± 2.59 & 90.05 ± 2.78 & 72.24 ± 11.33
\\
\rowcolor{gray!10} ConvMAE w/ ActiveSSF
& \textbf{98.73 ± 0.98} & \textbf{79.04 ± 4.79} & \textbf{73.67 ± 6.35} & \textbf{87.84 ± 0.88} & \textbf{75.52 ± 3.71} & \textbf{97.12 ± 0.51} & \textbf{73.54 ± 5.47} & \textbf{70.76 ± 8.21} & \textbf{96.72 ± 2.02} & \textbf{94.03 ± 1.50} & \textbf{81.12 ± 7.26}
\\
\hdashline
MIXMIM~\cite{Chen_2023_CVPR} w/o ActiveSSF
& 98.13 ± 1.09 & 68.13 ± 8.03 & 58.11 ± 4.26 & 79.69 ± 3.51 & 62.89 ± 7.04 & 95.08 ± 1.60 & 53.77 ± 8.98 & 56.03 ± 8.54 & 94.84 ± 1.52 & 79.32 ± 4.67 & 54.80 ± 19.83
\\
\rowcolor{gray!10} MIXMIM w/ ActiveSSF
& \textbf{98.28 ± 0.87} & \textbf{71.21 ± 2.76} & \textbf{63.17 ± 7.50} & \textbf{82.77 ± 1.62} & \textbf{67.79 ± 3.49} & \textbf{96.47 ± 0.98} & \textbf{58.26 ± 5.23} & \textbf{61.18 ± 10.81} & \textbf{96.41 ± 1.71} & \textbf{85.17 ± 2.44} & \textbf{70.54 ± 1.93}
\\
\hdashline
MAE~\cite{he2022masked} w/o ActiveSSF
& 98.53 ± 1.05 & 75.13 ± 5.13 & 70.54 ± 8.09 & 86.05 ± 0.59 & 74.25 ± 3.04 & 96.60 ± 0.72 & 67.66 ± 3.45 & 69.23 ± 7.68 & 97.00 ± 1.57 & 92.74 ± 2.77 & 77.20 ± 5.45
\\
\rowcolor{gray!10} 
MAE w/ ActiveSSF
& \textbf{98.63 ± 1.16} & \textbf{75.51 ± 6.56} & \textbf{74.20 ± 5.70} & \textbf{88.04 ± 0.71} & \textbf{76.51 ± 3.50} & \textbf{97.26 ± 0.74} & \textbf{73.17 ± 4.54} & \textbf{70.75 ± 8.23} & \textbf{97.68 ± 1.37} & \textbf{94.03 ± 1.67} & \textbf{78.79 ± 6.46}
\\
\hline\hline
\end{tabular}}}
\captionsetup{font=small}
\caption{
Performance comparison of self-supervised learning methods with and without ActiveSSF framework. Results are reported as PR-AUC scores (\%) for eleven megakaryocyte subtypes. $^{\dagger}$ indicates rare subtypes and $^{*}$ denotes common subtypes. Best results for each method pair are shown in bold.
}
\label{tab:compare_sota}    
\vspace{-10pt}
\end{table*}

\subsection{Datasets}
In this study, we assembled a comprehensive clinical dataset comprising stained blood cell slides specifically curated for megakaryocyte classification. The dataset includes 11 distinct subtypes of megakaryocytes. Small megakaryocyte with unlobed or minimally lobed nucleus (S-MK), Micromegakaryocyte (mMK), Multinucleated round megakaryocyte (MR-MK), and Large megakaryocyte with highly lobulated nucleus (HL-MK).

As shown in Fig.~\ref{fig:fig1}, our dataset exhibits a pronounced long-tailed distribution, with the G-MK subtype being dominant while CA-MK and HL-MK are notably underrepresented. In total, the dataset comprises 7,577 labeled images and 10,000 unlabeled blood cell slides, all acquired using professional medical scanning equipment. All cell annotations have been rigorously validated by experienced clinical experts to ensure high data quality. For experimental evaluation, we employed a five-fold cross-validation strategy to assess our method.

\begin{table}[htb]\small
    \belowrulesep=0pt
    \aboverulesep=0pt   
    \centering
    \caption{Comparison of different self-supervised learning methods with and without our proposed ActiveSSF framework. The symbol * indicates a p-value of less than 0.05 (paired t-test) when comparing with the baseline method. Best results for each method pair are shown in bold.}
    \resizebox{0.48\textwidth}{!}{
    \begin{tabular}{c|ccc}
        \toprule
        Method & Accuracy (\%) & PR-AUC (\%) & F1-score (\%)  \\
        \hline
        ResNet~\cite{he2016identity} & 76.15 ± 1.18 & 77.78 ± 1.77 & 71.97 ± 1.45 \\
        ViT~\cite{alexey2020image} & 68.45 ± 1.63 & 67.78 ± 1.68 &  60.03 ± 2.36 \\
        \hline
        MoCo v3~\cite{chen2021empirical} w/o ActiveSSF & 72.98 ± 0.68 & 73.31 ± 1.94 & 67.29 ± 1.71 \\
        MoCo v3 w/ ActiveSSF & \textbf{73.91 ± 1.13}$^{*}$ & \textbf{74.99 ± 1.73}$^{*}$ & \textbf{68.42 ± 1.50}$^{*}$ \\
        \hline
        ConvMAE~\cite{gao2022convmae} w/o ActiveSSF & 76.66 ± 0.59 & 79.72 ± 1.64 & 73.37 ± 1.36 \\
        ConvMAE w/ ActiveSSF & \textbf{81.10 ± 0.92}$^{*}$ & \textbf{84.37 ± 1.71}$^{*}$ & \textbf{78.85 ± 1.28}$^{*}$ \\
        \hline
        MIXMIM~\cite{Chen_2023_CVPR} w/o ActiveSSF & 72.52 ± 2.66 & 65.93 ± 4.48 & 72.80 ± 4.41 \\
        MIXMIM w/ ActiveSSF & \textbf{74.83 ± 0.67}$^{*}$ & \textbf{70.51 ± 0.68}$^{*}$ & \textbf{77.39 ± 1.34}$^{*}$ \\
        \hline
        MAE~\cite{he2022masked} w/o ActiveSSF & 78.76 ± 1.06 & 82.27 ± 1.35 & 76.34 ± 1.31 \\
        MAE w/ ActiveSSF & \textbf{80.36 ± 1.03}$^{*}$ & \textbf{84.05 ± 1.37}$^{*}$ & \textbf{77.64 ± 1.47}$^{*}$ \\
        \bottomrule
    \end{tabular}
    }
    \vspace{0cm}
    \label{tab:exp}
\end{table}

\begin{algorithm}[h]
\caption{ActiveSSF Framework}
\label{alg:activessf}
\SetAlgoLined
\SetNoFillComment
\SetArgSty{textnormal}
\small{\KwIn{Labeled dataset $T$, Unlabeled dataset $U$, Gaussian kernel $\sigma=1.5$, Scaling factor $\alpha=0.5$, Fill rate threshold $\tau=0.7$}}
\small{\KwOut{Selected samples $S'$ for self-supervised learning}}

\BlankLine
{\footnotesize{\color{DarkBlue}{\tcc{Cell Region Filtering Stage}}}}

\For{each image $D_i \in U$}{
    Apply Gaussian filtering: $G_i = D_i * G(x,y;\sigma)$
    
    Two-stage K-means clustering: $20 \rightarrow 10$ clusters
    
    Transform to HSV: $H_i = \text{RGB2HSV}(G_i)$
    
    Generate masks: $M_i = \text{HSVThreshold}(H_i)$
    
    Extract regions: $S_i = \{r : \text{FillRate}(r) \geq \tau\}$
}

\BlankLine
{\footnotesize{\color{DarkBlue}{\tcc{Prototype Generation Stage}}}}

Extract features: $F = \text{ResNet}(T)$

Initialize cluster centers $\{c_1,...,c_K\}$

\While{not converged}{
    Update assignments: $C_i$ via~\Cref{eq:cluster}
    
    Update centers: $c_i$ via~\Cref{eq:cluster1}
}

\BlankLine
{\footnotesize{\color{DarkBlue}{\tcc{Active Sample Selection Stage}}}}

\For{each cluster $i$}{

    Compute $\text{threshold}_i$ via~\Cref{eq:threshold}
    
    \For{each region in $S_i$}{
        $x = \text{ResNet}(\text{region})$
        
        \If{$\|x-c_i\|_2 \leq \text{threshold}_i$}{
            $S'_i = S'_i \cup \{\text{region}\}$
        }
    }
}

\Return{$S' = \{S'_1,...,S'_K\}$}
\end{algorithm}

\subsection{Implementation Details}

The proposed method was implemented using the PyTorch framework and evaluated on NVIDIA RTX 3090 GPUs. For the cell region filtering stage, we processed 10,000 unlabeled megakaryocyte images utilizing a \(3\times3\) Gaussian kernel with \(\sigma = 1.5\), which resulted in 7,299 filtered samples.

For model training, we employed the AdamW optimizer (with a weight decay of 0.05) and a batch size of 64. The learning rate was set to \(1\times10^{-3}\times (\text{batch size}/256)\), with a warmup phase lasting 5 epochs. The model was first pre-trained for 400 epochs, followed by 50 epochs of fine-tuning. During both the pre-training and fine-tuning stages, a layer-wise learning rate decay of 0.75 was maintained.

\begin{figure}[!ht]
    \centering
    \begin{minipage}[b]{0.32\linewidth}
        \centering
        \includegraphics[width=\linewidth]{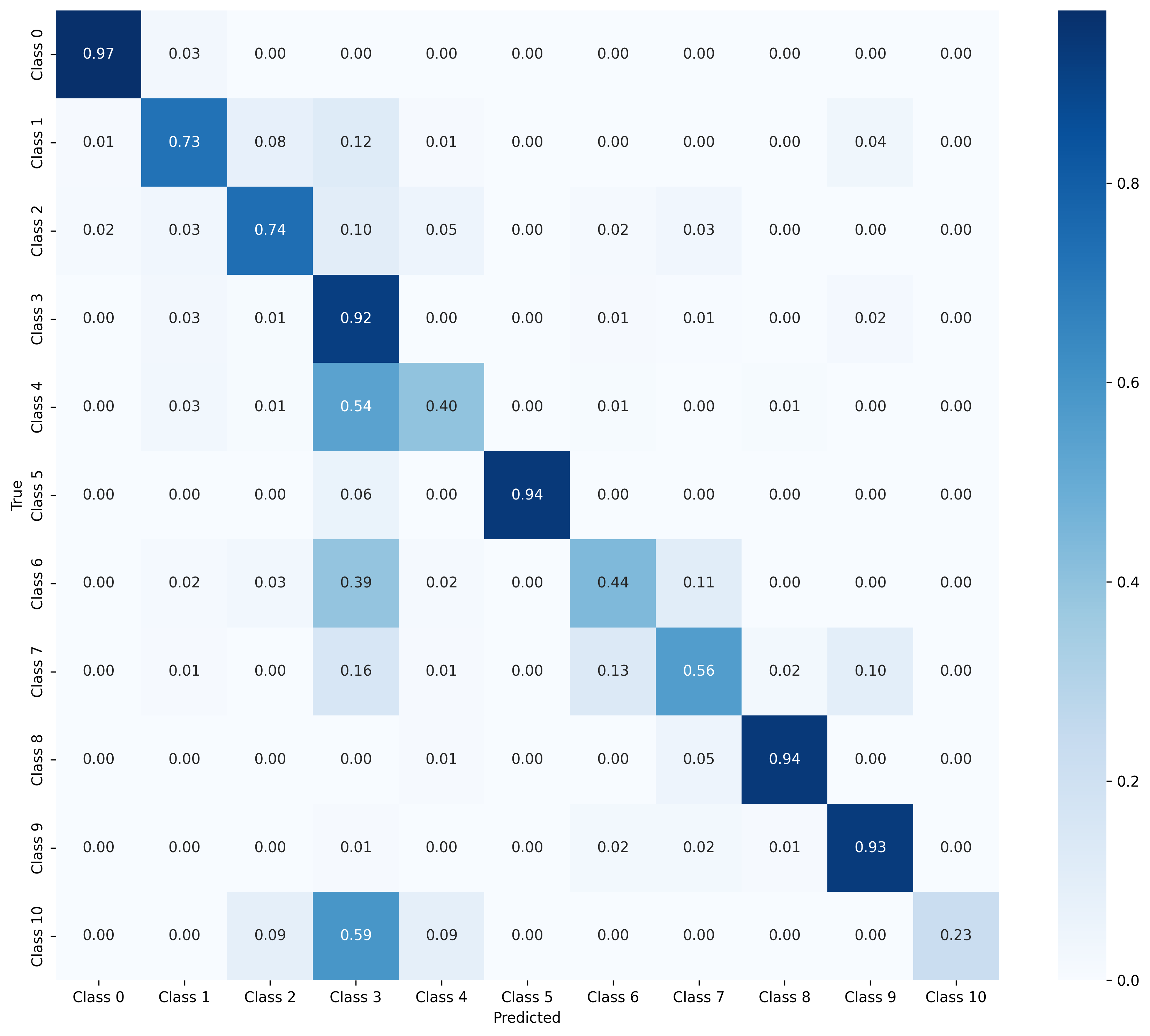}
        \centerline{(a)}
    \end{minipage}
    \hfill
    \begin{minipage}[b]{0.32\linewidth}
        \centering
        \includegraphics[width=\linewidth]{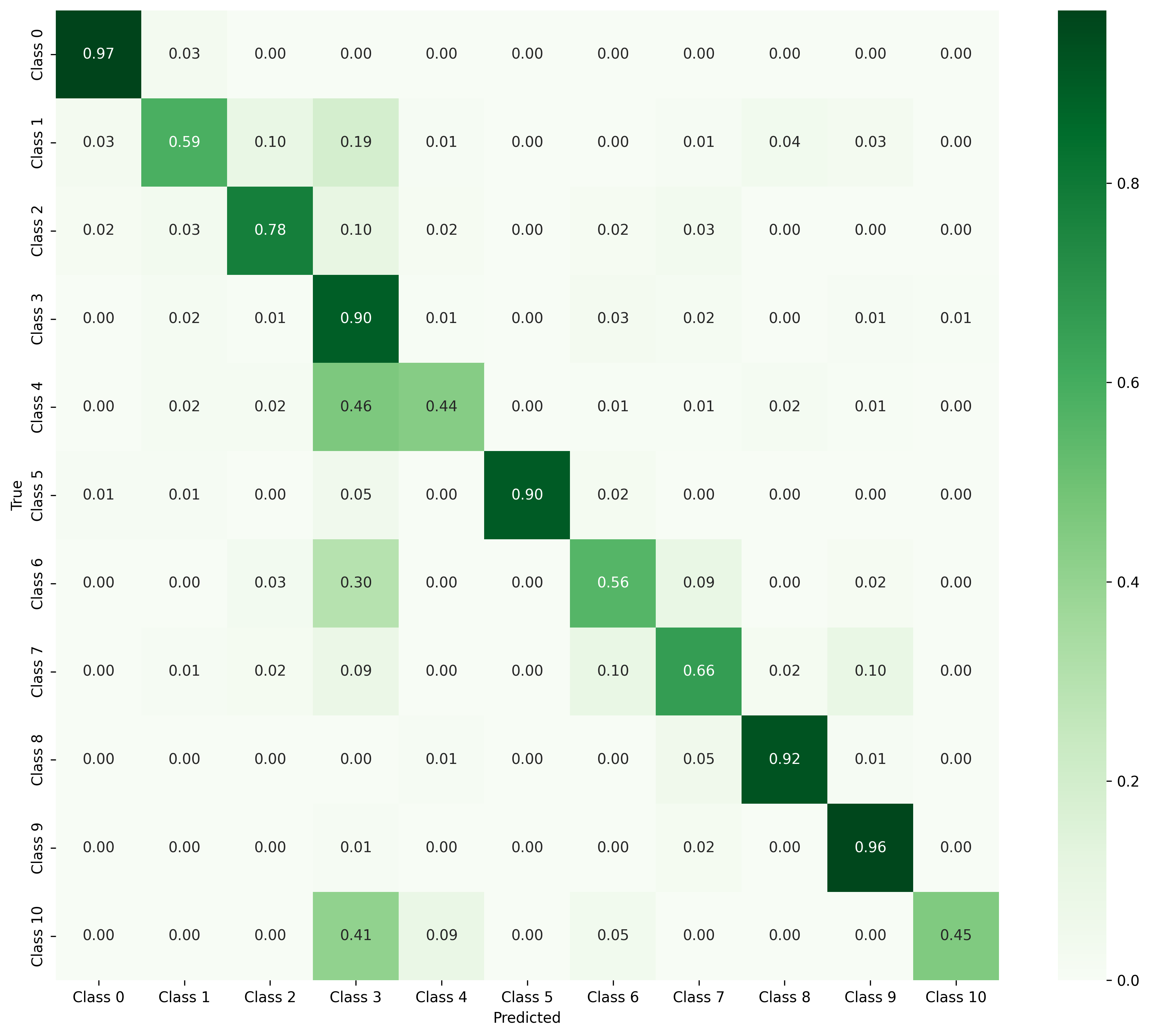}
        \centerline{(b) }
    \end{minipage}
    \hfill
    \begin{minipage}[b]{0.32\linewidth}
        \centering
        \includegraphics[width=\linewidth]{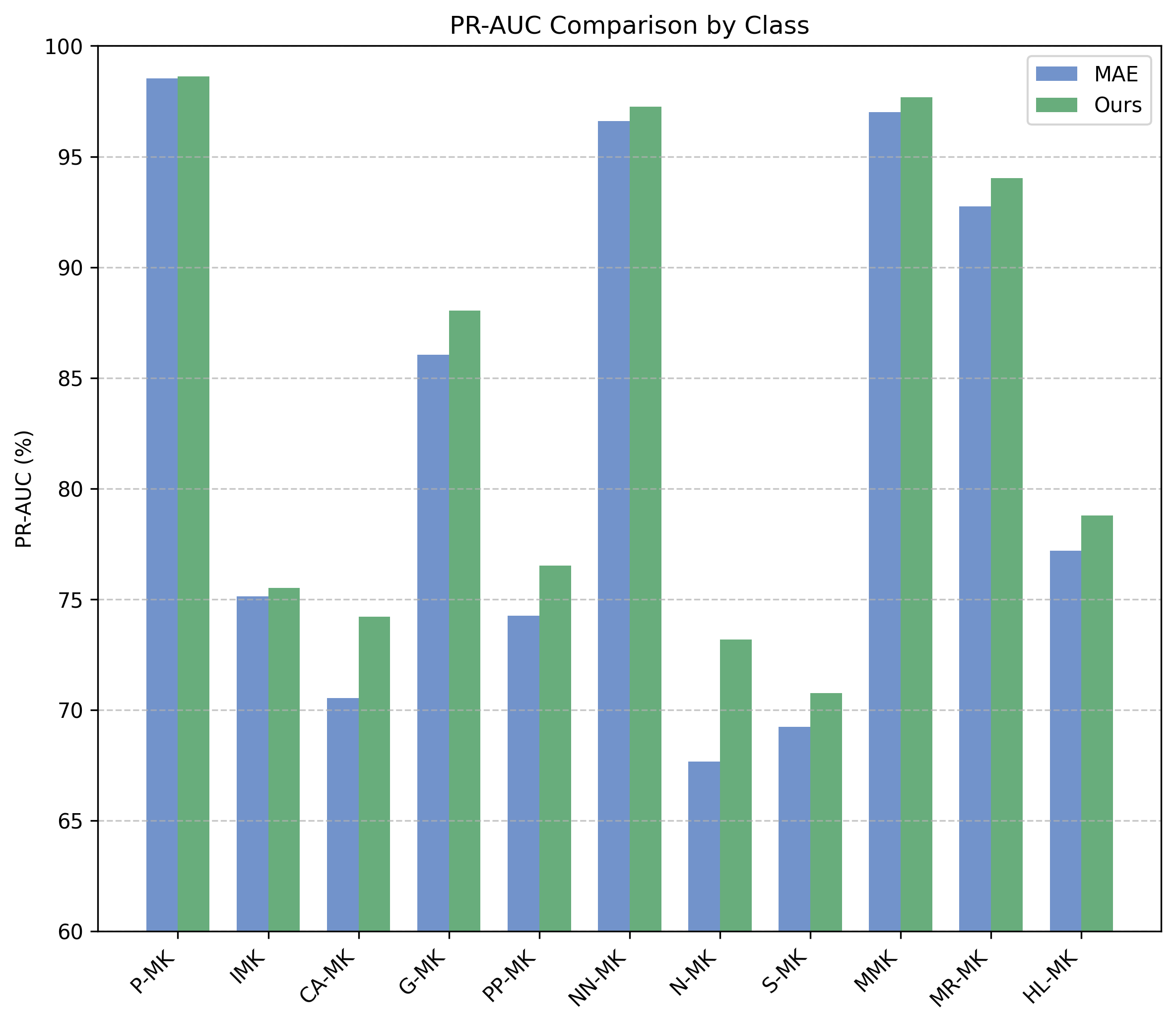}
        \centerline{(c)}
    \end{minipage}
    
    \caption{Comparison of classification performance across all categories, between MAE and MAE+ActiveSSF: (a) MAE Confusion Matrix, (b) MAE+ActiveSSF Confusion Matrix, and (c) Per-Class PR-AUC score Comparison.}
    \label{fig:result2}
    \vspace{-15px}
\end{figure}

We evaluated our approach using various backbone networks including ResNet~\cite{he2016identity}, ViT~\cite{alexey2020image}, and SOTA self-supervised models~\cite{chen2021empirical,gao2022convmae,he2022masked}. The dynamic threshold used in sample selection was set to \(17 \pm 2\) with a temperature parameter \(\alpha\) of 1. To ensure a fair comparison, all models were fine-tuned on the same 11-class classification task with a uniform input size of \(224 \times 224\). All experiments were conducted using five-fold cross-validation, and statistical significance was assessed using paired t-tests (all p-values \(< 0.05\)).

\subsection{Results}

\noindent \textbf{Comparison with State-of-the-Art.} We comprehensively evaluated our ActiveSSF framework against state-of-the-art methods in megakaryocyte classification. The baseline methods include traditional architectures (ResNet~\cite{he2016identity}, ViT~\cite{alexey2020image}) and recent self-supervised learning approaches (MAE~\cite{he2022masked}, ConvMAE~\cite{gao2022convmae}, MoCo v3~\cite{chen2021empirical}, and MIXMIM~\cite{Chen_2023_CVPR}).

As shown in Table~\ref{tab:exp}, ActiveSSF consistently improves performance across various self-supervised models. Notably, when integrated with ConvMAE, our framework achieves the best overall performance.

Furthermore, as detailed in Table~\ref{tab:compare_sota}, we evaluated ActiveSSF against several SOTA self-supervised learning methods across the 11 megakaryocyte subtypes. The results demonstrate that ActiveSSF consistently outperforms existing approaches across different backbone networks. Of particular note, our method yields significant improvements in classifying rare subtypes.   

Moreover, our framework demonstrates a superior capability in addressing long-tailed distributions, as visualized in Figure~\ref{fig:result2}. The performance improvements are particularly pronounced for rare subtypes, indicating that ActiveSSF effectively mitigates the class imbalance challenge in megakaryocyte classification.

\begin{table}[t]
    \belowrulesep=0pt
    \aboverulesep=0pt   
    \centering
    \caption{Ablation study of ActiveSSF components with MAE on the megakaryocyte dataset. All experiments were conducted using five-fold cross-validation to ensure reliability. Best results are shown in bold.}
    \resizebox{0.48\textwidth}{!}{
    \begin{tabular}{cccc|ccc}
        \toprule
        ViT~\cite{alexey2020image} & MAE~\cite{he2022masked} & CRF & ASS & Accuracy (\%) & PR-AUC (\%) & F1-score (\%) \\
        \hline
        \checkmark & & & & 68.45 ± 1.63 & 67.78 ± 1.68 &  60.03 ± 2.36 \\
        \checkmark & \checkmark & & & 78.76 ± 1.06 & 82.27 ± 1.35 & 76.34 ± 1.31 \\
        \checkmark & \checkmark & \checkmark & & 79.79 ± 1.08 &  83.80 ± 1.59 & 76.71 ± 2.03 \\
        \hline
        \checkmark & \checkmark & \checkmark & \checkmark & \textbf{80.36 ± 1.03} & \textbf{84.05 ± 1.37} & \textbf{77.64 ± 1.47} \\
        \bottomrule
    \end{tabular}
    } 
    \vspace{0cm}
    \label{tab:Ablation}
    \vspace{-15px}
\end{table}

\noindent \textbf{Ablation Study.}  We conducted a series of ablation experiments to evaluate the contribution of each module within our framework. Specifically, we compared four configurations: (1) using the Vision Transformer (ViT)~\cite{alexey2020image} as the baseline without any additional modules; (2) incorporating MAE~\cite{he2022masked} pretraining; (3) further applying Cell Region Filtering (CRF); and (4) finally, integrating Active Sample Selection (ASS), where similarity thresholds are dynamically adjusted based on sample density information to enhance performance on long-tailed distributions. Table~\ref{tab:Ablation} summarizes the quantitative results, clearly demonstrating the specific contributions of each module to the overall performance.

\section{Conclusion and Discussion} 
\label{sec:majhead}

In this paper, we introduced ActiveSSF, a novel framework that integrates active learning strategies with self-supervised pretraining for the classification of megakaryocytes. By combining clinical prior knowledge-based cell region filtering, adaptive sample selection, and prototype clustering on labeled samples, our approach effectively addresses three critical challenges: (1) suppressing background noise to reveal informative cellular regions, (2) managing long-tailed class distributions through dynamic threshold adjustment, and (3) mitigating the impact of complex morphological variations by generating representative feature prototypes. Extensive experiments on clinical datasets demonstrate that ActiveSSF achieves significant improvements over state-of-the-art methods, particularly enhancing the recognition accuracy of rare subtypes. These promising results suggest that ActiveSSF provides a robust solution for automated blood cell analysis in clinical settings. Future work will explore extending this framework to other medical imaging domains characterized by scarce labeled data and inherently imbalanced class distributions.

\section{Ethics Statement}

The experimental procedures involving human subjects described in this paper were approved by the Clinical Medical Ethics Committee of Xiangya Hospital, Central South University.



\bibliographystyle{IEEEtran} 
\bibliography{refs}

\begin{thebibliography}{10}
\providecommand{\url}[1]{#1}
\csname url@samestyle\endcsname
\providecommand{\newblock}{\relax}
\providecommand{\bibinfo}[2]{#2}
\providecommand{\BIBentrySTDinterwordspacing}{\spaceskip=0pt\relax}
\providecommand{\BIBentryALTinterwordstretchfactor}{4}
\providecommand{\BIBentryALTinterwordspacing}{\spaceskip=\fontdimen2\font plus
\BIBentryALTinterwordstretchfactor\fontdimen3\font minus \fontdimen4\font\relax}
\providecommand{\BIBforeignlanguage}[2]{{%
\expandafter\ifx\csname l@#1\endcsname\relax
\typeout{** WARNING: IEEEtran.bst: No hyphenation pattern has been}%
\typeout{** loaded for the language `#1'. Using the pattern for}%
\typeout{** the default language instead.}%
\else
\language=\csname l@#1\endcsname
\fi
#2}}
\providecommand{\BIBdecl}{\relax}
\BIBdecl

\bibitem{he2016identity}
K.~He, X.~Zhang, S.~Ren, and J.~Sun, ``Identity mappings in deep residual networks,'' in \emph{Computer Vision--ECCV 2016: 14th European Conference, Amsterdam, The Netherlands, October 11--14, 2016, Proceedings, Part IV 14}.\hskip 1em plus 0.5em minus 0.4em\relax Springer, 2016, pp. 630--645.

\bibitem{macqueen1967some}
J.~MacQueen, ``Some methods for classification and analysis of multivariate observations,'' in \emph{Proceedings of 5-th Berkeley Symposium on Mathematical Statistics and Probability/University of California Press}, 1967.

\bibitem{su2014neural}
M.-C. Su, C.-Y. Cheng, and P.-C. Wang, ``A neural-network-based approach to white blood cell classification,'' \emph{The scientific world journal}, vol. 2014, no.~1, p. 796371, 2014.

\bibitem{TOMARI2014206}
\BIBentryALTinterwordspacing
R.~Tomari, W.~N.~W. Zakaria, M.~M.~A. Jamil, F.~M. Nor, and N.~F.~N. Fuad, ``Computer aided system for red blood cell classification in blood smear image,'' \emph{Procedia Computer Science}, vol.~42, pp. 206--213, 2014, medical and Rehabilitation Robotics and Instrumentation (MRRI2013). [Online]. Available: \url{https://www.sciencedirect.com/science/article/pii/S1877050914014914}
\BIBentrySTDinterwordspacing

\bibitem{dong2022fusing}
X.~Dong, M.~Li, P.~Zhou, X.~Deng, S.~Li, X.~Zhao, Y.~Wu, J.~Qin, and W.~Guo, ``Fusing pre-trained convolutional neural networks features for multi-differentiated subtypes of liver cancer on histopathological images,'' \emph{BMC Medical Informatics and Decision Making}, vol.~22, no.~1, p. 122, 2022.

\bibitem{10508987}
X.~Zhao, Z.~Li, X.~Luo, P.~Li, P.~Huang, J.~Zhu, Y.~Liu, J.~Zhu, M.~Yang, S.~Chang, and J.~Dong, ``Ultrasound nodule segmentation using asymmetric learning with simple clinical annotation,'' \emph{IEEE Transactions on Circuits and Systems for Video Technology}, pp. 1--1, 2024.

\bibitem{10635682}
X.~Zhao, P.~Li, X.~Luo, M.~Yang, S.~Chang, and Z.~Li, ``Sam-driven weakly supervised nodule segmentation with uncertainty-aware cross teaching,'' in \emph{2024 IEEE International Symposium on Biomedical Imaging (ISBI)}, 2024, pp. 1--5.

\bibitem{zhao2024hfgs}
H.~Zhao, X.~Zhao, L.~Zhu, W.~Zheng, and Y.~Xu, ``Hfgs: 4d gaussian splatting with emphasis on spatial and temporal high-frequency components for endoscopic scene reconstruction,'' \emph{arXiv preprint arXiv:2405.17872}, 2024.

\bibitem{li2023d}
Z.~Li, Z.~Shang, J.~Liu, H.~Zhen, E.~Zhu, S.~Zhong, R.~N. Sturgess, Y.~Zhou, X.~Hu, X.~Zhao \emph{et~al.}, ``D-lmbmap: a fully automated deep-learning pipeline for whole-brain profiling of neural circuitry,'' \emph{Nature Methods}, vol.~20, no.~10, pp. 1593--1604, 2023.

\bibitem{chen2021empirical}
X.~Chen, S.~Xie, and K.~He, ``An empirical study of training self-supervised vision transformers,'' in \emph{Proceedings of the IEEE/CVF international conference on computer vision}, 2021, pp. 9640--9649.

\bibitem{gao2022convmae}
P.~Gao, T.~Ma, H.~Li, Z.~Lin, J.~Dai, and Y.~Qiao, ``Convmae: Masked convolution meets masked autoencoders,'' \emph{arXiv preprint arXiv:2205.03892}, 2022.

\bibitem{Chen_2023_CVPR}
K.~Chen, Z.~Liu, L.~Hong, H.~Xu, Z.~Li, and D.-Y. Yeung, ``Mixed autoencoder for self-supervised visual representation learning,'' in \emph{Proceedings of the IEEE/CVF Conference on Computer Vision and Pattern Recognition (CVPR)}, June 2023, pp. 22\,742--22\,751.

\bibitem{he2022masked}
K.~He, X.~Chen, S.~Xie, Y.~Li, P.~Doll{\'a}r, and R.~Girshick, ``Masked autoencoders are scalable vision learners,'' in \emph{Proceedings of the IEEE/CVF conference on computer vision and pattern recognition}, 2022, pp. 16\,000--16\,009.

\bibitem{he2020momentum}
K.~He, H.~Fan, Y.~Wu, S.~Xie, and R.~Girshick, ``Momentum contrast for unsupervised visual representation learning,'' in \emph{Proceedings of the IEEE/CVF conference on computer vision and pattern recognition}, 2020, pp. 9729--9738.

\bibitem{alexey2020image}
D.~Alexey, ``An image is worth 16x16 words: Transformers for image recognition at scale,'' \emph{arXiv preprint arXiv: 2010.11929}, 2020.

\end{thebibliography}

\end{document}